\PassOptionsToPackage{dvipsnames,table}{xcolor}
\documentclass[10pt,twocolumn,letterpaper]{article}

\usepackage[pagenumbers]{cvpr} 

\newcommand{\topic}[1]
{
\vspace{1mm}\noindent\textbf{#1}
}

\DeclareMathAlphabet{\altmathcal}{OMS}{cmsy}{m}{n}
\DeclareMathAlphabet{\mathbfit}{OT1}{ptm}{bx}{it}

\usepackage{soul} 
\usepackage{multirow}
\usepackage{amsmath, amssymb}
\usepackage{pifont}
\usepackage{enumitem, array}
\newcolumntype{C}[1]{>{\centering\arraybackslash}p{#1}}
\newcolumntype{L}[1]{>{\raggedright\arraybackslash}p{#1}}
\newcolumntype{R}[1]{>{\raggedleft\arraybackslash}m{#1}}

\usepackage{blindtext}
\usepackage{graphicx}
\usepackage{caption}

\usepackage[symbol]{footmisc}
\def\D{\mathcal{D}}  
\def\I{\mathcal{I}}  
\def\V{\mathcal{V}} 
\def\m{M} 
\def\M{\mathcal{M}} 
\def\pose{\pi}     

\def\diffz{\mathbf{z}} 
\def\difft{t_d} 
\def\diffT{T_d} 
\def\diffunet{\Phi} 
\def\diffc{\mathbf{c}} 
\def\rev{\overline}
\def\trweight{\gamma} 
\def\trstep{{\tau_{\mathrm{tr}}}}
\def\refinestep{{\tau_{\mathrm{refine}}}}

\def\C{\mathcal{C}}  
\def\gscolor{c} 
\def\a{\alpha}       
\def\gspos{\mu} 
\def\gsmlp{\Theta} 
\def\hex{\mathcal{H}} 
\def\embed{\mathbf{w}} 

\def\R{\mathbb{R}} 

\def\prompt{\mathtt{p}} 
\def\pcd{\mathcal{P}} 

\def\ray{\mathbf{v}} 
\def\lookat{\mathbf{l}} 

\def\Loss{\mathcal{L}} 

\definecolor{cvprblue}{rgb}{0.21,0.49,0.74}
\usepackage[pagebackref,breaklinks,colorlinks,citecolor=cvprblue]{hyperref}

\title{VividDream: Generating 3D Scene with Ambient Dynamics}

\author{\vspace{2mm}
Yao-Chih Lee$^{1}$
\hspace{-0.05cm}
Yi-Ting Chen$^{1}$
\hspace{-0.05cm}
Andrew Wang$^{1}$
\hspace{-0.05cm}
Ting-Hsuan Liao$^{1}$
\hspace{-0.05cm}
Brandon Y. Feng$^{2}$
\hspace{-0.05cm}
Jia-Bin Huang$^{1}$
\\ 
$^{1}$University of Maryland, College Park
\hspace{0.5cm}
$^{2}$Massachusetts Institute of Technology
\\ \vspace{2mm}
\url{https://vivid-dream-4d.github.io}
\vspace{-1cm}
}

\begin{document}
\twocolumn[{
\renewcommand\twocolumn[1][]{#1}
\maketitle
\begin{center}
    \centering
    \captionsetup{type=figure}
    \includegraphics[trim={0cm 21cm 31.3cm 0cm},clip,width=\textwidth]{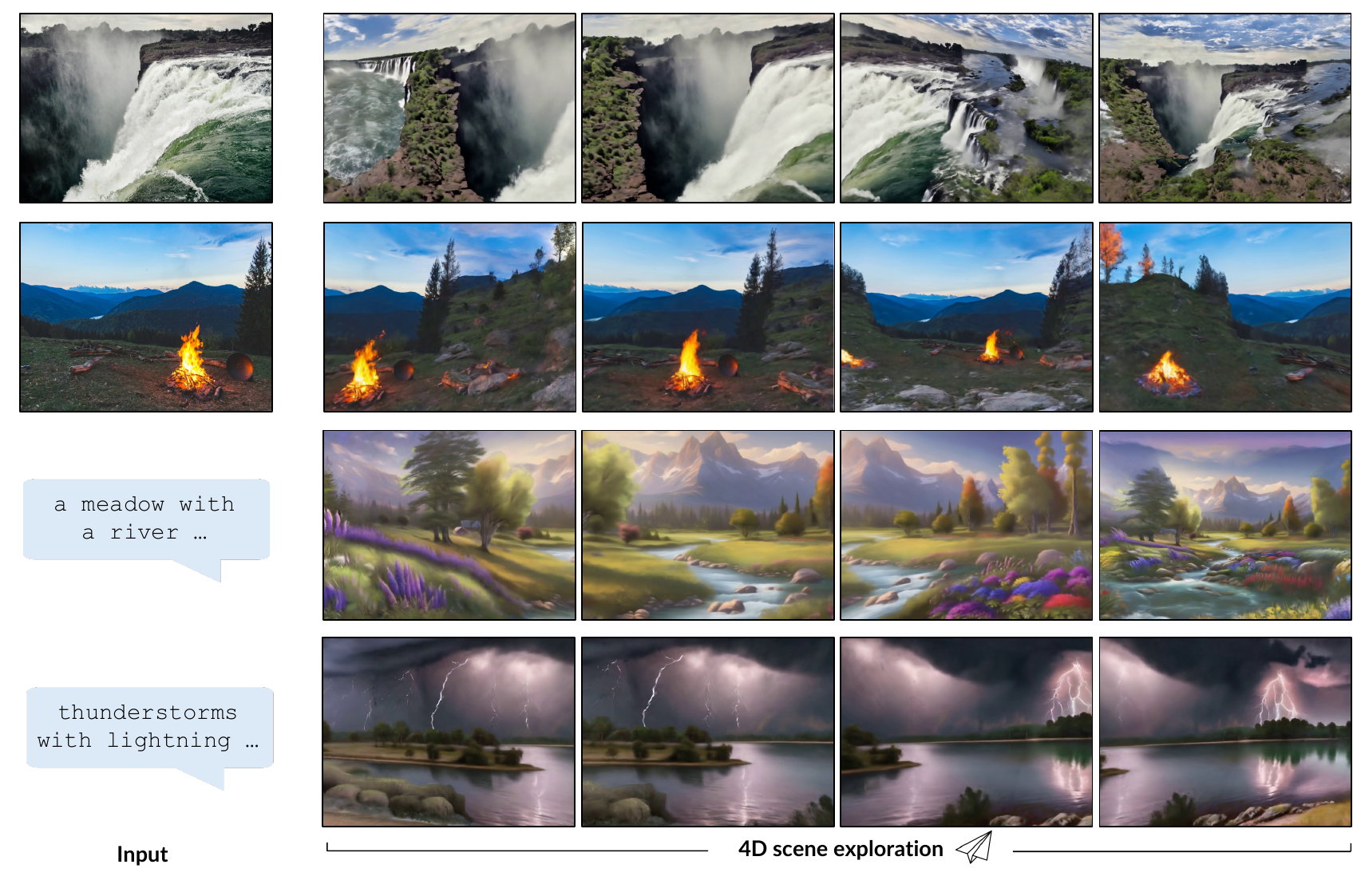}
    \vspace{-10mm}
    \captionof{figure}{
    \textbf{Generating 3D Scene with Ambient Dynamics.} Given an input image or text prompt, our proposed method, VividDream, generates a large, explorable 4D scene with ambient scene motion.
    }
    \label{fig:teaser}
\end{center}
}]

\maketitle

\begin{abstract}
We introduce VividDream, a method for generating explorable 4D scenes with ambient dynamics from a single input image or text prompt. 
VividDream first expands an input image into a static 3D point cloud through iterative inpainting and geometry merging. 
An ensemble of animated videos is then generated using video diffusion models with quality refinement techniques and conditioned on renderings of the static 3D scene from the sampled camera trajectories.
We then optimize a canonical 4D scene representation using an animated video ensemble, with per-video motion embeddings and visibility masks to mitigate inconsistencies. 
The resulting 4D scene enables free-view exploration of a 3D scene with plausible ambient scene dynamics. 
Experiments demonstrate that VividDream can provide human viewers with compelling 4D experiences generated based on diverse real images and text prompts. 
\end{abstract}
\section{Introduction}

Recent advancements in text-to-image generation~\cite{dhariwal2021diffusion, nichol2021improved, rombach2022high} have revolutionized the field of computer vision, producing highly realistic and contextually accurate images from textual descriptions. 
This progress has paved the way for extending generative models beyond static images to higher-dimensional outputs, including video generation, 3D object generation, and 3D scene generation.

However, in the realm of 4D space, current research predominantly focuses on generating or reconstructing individual 4D objects~\cite{bahmani20234d, zheng2023unified,ren2023dreamgaussian4d, li2024loopgaussian, zhang2024physdreamer, zeng2024stag4d}. While these efforts have led to significant breakthroughs in capturing objects' dynamics over time, there remains a notable gap in the generation of comprehensive 4D scenes. 
A 4D scene encompasses not just the temporal evolution of the scene but also the spatial scale of the 3D environment, enabling immersive view exploration.
Although the recent DreamScene4D~\cite{chu2024dreamscene4d} generates a 4D scene from an input video, it focuses on reconstructing and generating the dynamic foreground objects but not expanding the 3D scene from the input views for larger view exploration.


\begin{figure}
\centering
\includegraphics[trim={1.2cm 30.5cm 39cm 0},clip,width=0.97\linewidth]{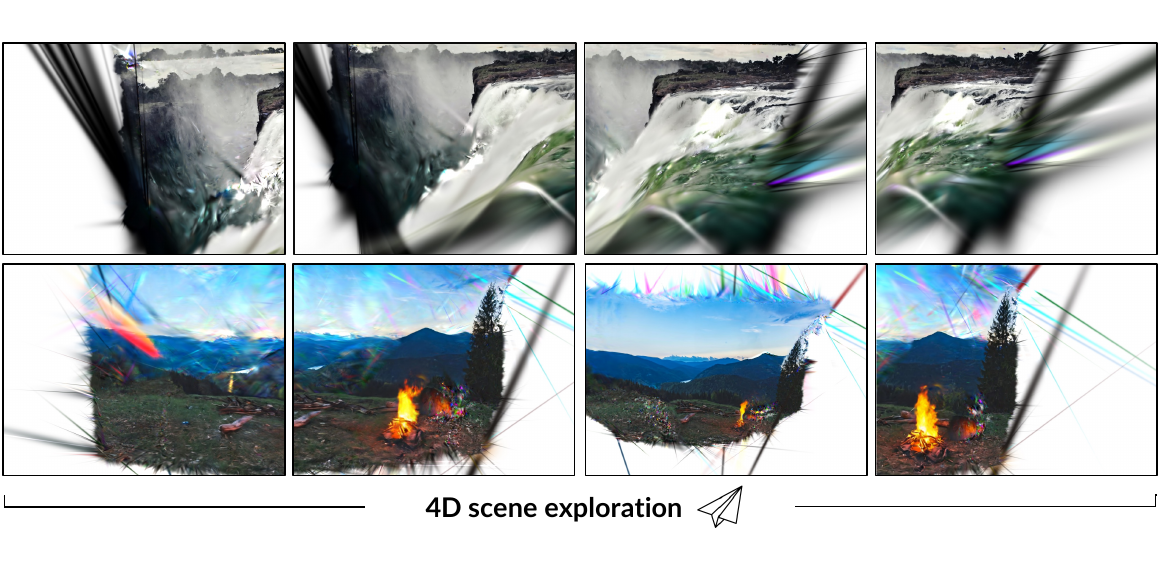}
\vspace{-6mm}
\caption{\textbf{Failure in a baseline solution.} 
Training a 4D scene representation (e.g., 4DGS~\cite{4dgs}) using a single video generated by SVD~\cite{svd}  poses several challenges. 
First, the weak or absent camera motion in the animated video leads to many unseen areas and limited novel view synthesis. 
Second, existing video depth and pose estimation algorithms~\cite{casualsam,kopf2021robust} have difficulty in handling scenes with ambient motion due to their reliance on accurate correspondence estimation (\,  e.g., optical flow~\cite{raft}). 
These limitations motivate our multi-video approach to 4D scene generation.
}\vspace{-6mm}
\label{fig:baseline_failure}
\end{figure}

In this paper, we aim to generate 4D scenes, enabling users to explore the 3D scene with dynamic ambient motion, such as fluid motions, trees, and grass swaying in the wind. 
To achieve this, a straightforward baseline might involve using a powerful video generator to create a sequence and then performing 3D reconstruction~\cite{kopf2021robust,casualsam} on this video to form a 4D scene~\cite{liu2023robust,li2023dynibar,4dgs,lee2023fast}. However, this approach falls short in several ways. 
No matter how detailed, a single video is insufficient to capture the full complexity and navigability required for an explorable 4D scene (Fig.~\ref{fig:baseline_failure}). 
Such a scene necessitates multiple perspectives and the ability to transition between different views and moments seamlessly.

To address these limitations, we propose a novel pipeline for generating explorable 4D scenes from a single image or text prompt.
In Stage 1, we expand the initial 3D point cloud obtained from the input through an iterative view extrapolation and inpainting process. 
Stage 2 focuses on generating ambient scene motion by animating multiple view-extrapolation videos rendered from the 3D scene. 
Finally, in Stage 3, we train a 4D scene representation using the animated multi-view videos while handling inconsistencies through visibility masking and per-video motion embeddings.

Our key contributions include:
\begin{itemize}
    \item A holistic approach to generating explorable 4D scenes with ambient dynamics from a single image or text prompt.
    \item A multi-video animation framework to generate scene motion while respecting specified camera trajectories.
    \item Techniques to mitigate inconsistencies when generating a 4D scene from multiple independently animated videos.
\end{itemize}

We demonstrate the effectiveness of our approach in a variety of scenes. 
Experimental results showcase our method's ability to generate compelling 4D scene experiences with plausible ambient dynamics.
We will release source code to facilitate future research on this problem.
\section{Related Work}
\paragraph{Text-to-3D generation}
Generative models have achieved promising results in generating realistic 3D objects~\cite{poole2022dreamfusion, chen2023fantasia3d, lin2023magic3d, wang2024prolificdreamer, metzer2023latent,tang2023dreamgaussian} and
3D scenes.
By leveraging the power of 2D diffusion models to generate high-quality 2D images, certain approaches~\cite{hollein2023text2room, fridman2024scenescape, yu2023wonderjourney} address text-to-3D scene generation as an inpainting problem.
These methods take text prompts as input and utilize 2D diffusion models~\cite{rombach2022high,lugmayr2022repaint,alzayer2024magicfixup} to generate images and inpaint the parts not seen in previous images. 
Another trend seeks higher controllability over the generated content by pre-defining the allocation of objects within the scene~\cite{zhou2024gala3d, zhang2023scenewiz3d, cohen2023set}. 
However, these methods generate static 3D scenes, lacking dynamic ambient scene motion, essential for better immersion. 
VividDream aims to generate dynamic scenes with ambient dynamics to provide a more immersive and engaging experience.

\paragraph{Video diffusion models}
Recent advancements in video diffusion models have demonstrated significant progress in generating high-quality video content~\cite{luo2023videofusion,blattmann2023align,ge2023preserve,xing2023dynamicrafter,svd}. Text-guided video diffusion models~\cite{ho2022imagen,singer2022make,blattmann2023align,ge2023preserve,wang2023lavie} extend the capabilities of the powerful text-to-image generators~\cite{imagen,rombach2022high} to the domain of video generation. To enhance the context guidance, some methods condition the video generation with images~\cite{chen2023videocrafter1,xing2023dynamicrafter,svd,guo2023animatediff}.  
However, the text and/or image conditions can still be insufficient to fully control the generated motions and separate camera motion from scene motion. 
To improve the controllability of motions, MotionCtrl~\cite{wang2023motionctrl} and CameraCtrl~\cite{he2024cameractrl} fine-tune pre-trained video diffusion models~\cite{svd,guo2023animatediff} with additional conditioning modules to take the user-specified camera or scene motion guidance. On the other hand, Time-Reversal~\cite{timereversal} leverages Stable Video Diffusion (SVD)~\cite{svd} in a zero-shot manner to guide video generation with an end-view condition and can achieve view interpolation and video looping. We build on the camera controllability introduced by Time-Reversal and focus on generating ambient scene motion with conditioned camera poses. 

\paragraph{Single-image animation.}
Cinemagraph generation~\cite{holynski2021animating,mahapatra2022controllable,fan2023simulating} animates a single image into a video with looping motions. Text2-Cinemagraph~\cite{mahapatra2023text} enables text-guided animation of the generated images. Li~\etal~\cite{li2023generative} predicts the ambient scene motion and provides the interactivity on the generated dynamics.
Li~\etal~\cite{li2023_3dcinemagraphy} lifts the 2D cinemagraph into 3D space to allow small 3D viewpoint changes. 
Nevertheless, these methods are still limited in the input image view without expanding the scene. In contrast, we expand a 4D scene from an input image to allow exploration with significant viewpoint change.

\begin{figure*}
\centering
\includegraphics[trim={0cm 15cm 28cm 0cm},clip,width=0.95\linewidth]{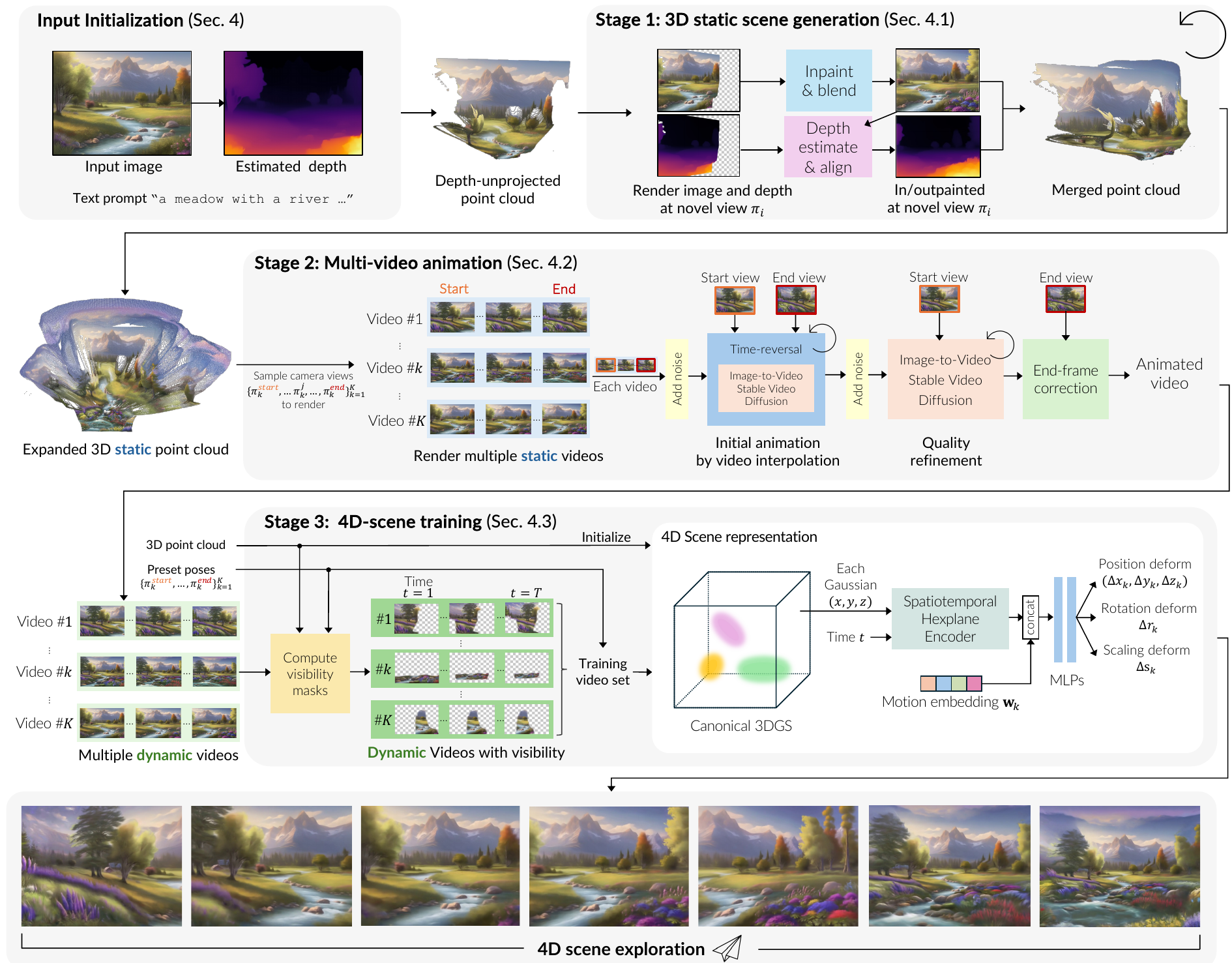}
\vspace{-4.0mm} 
\caption{\textbf{Method overview.} 
Our method takes either a single image or a text prompt as input. 
For an image input, we generate a caption using~\cite{blip2} and estimate its depth using DepthAnything~\cite{depthanything}. 
For a text prompt input, we generate the corresponding image using a text-to-image generator~\cite{rombach2022high}. 
The image, text prompt, and estimated depth form the initialization for the subsequent stages.
In Stage~1 (Sec.~\ref{sec:3dscenegen}), we expand the initial 3D point cloud through an iterative process of novel view inpainting and point cloud merging using aligned depth estimates. 
Stage~2 (Sec.~\ref{sec:video_gen}) focuses on generating the scene motions by rendering $K$ static view-extrapolation videos covering the entire 3D scene. 
Each video is first animated using Time-Reversal~\cite{timereversal} with static renderings as conditions.
To improve the video quality, we refine the animated videos using SVD~\cite{svd}.
However, this refinement may cause the camera motion to deviate from the desired trajectory. We mitigate this issue by applying a smooth transition on the last few frames using FILM~\cite{film} to match the conditioning end view.
Finally, in Stage~3 (Sec.~\ref{sec:train4d}), we train the 4D scene model, 4DGS~\cite{4dgs}, using the animated videos from Stage 2. 
To handle appearance and motion inconsistencies among the multi-view videos, we apply visibility masks with soft blending weights and introduce a per-video motion embedding inspired by NeRF-W~\cite{nerfw}. 
The resulting 4D scene model enables consistent motion and immersive 4D scene exploration.
}\vspace{-3mm}
\label{fig:pipeline}
 \end{figure*}

\section{Preliminary}
\subsection{SVD for image-to-video generation}
SVD~\cite{svd} is a publicly available state-of-the-art video generator that builds upon Stable Diffusion~\cite{rombach2022high}, producing 14 or 25 frames at a resolution of 1024$\times$ 768. 
It takes in an image and generates a video sequence by a latent-based diffusion model, where a 3D-UNet, $\diffunet$, denoises a video latent $\diffz_{\diffT}\sim \mathcal{N}(\mathbf{0}, \mathbf{I})$ into a clean $\diffz_{0}$ over $\diffT$ steps.
At each denoising step $\difft$, the 3D-UNet $\diffunet$ denoises the video latent $\diffz_{\difft}$ by $    \diffz_{\difft - 1} = \diffunet(\diffz_{\difft} \oplus \diffz^{\mathrm{img}}, \difft, \diffc^{\mathrm{img}}, \diffc^{\mathrm{scalar}})$,
where $\oplus$ denotes the concatenating operation, $\diffz^{\mathrm{img}}$ is the input image condition encoded by VAE. 
Additionally, $\diffc^{\mathrm{img}}$ and $\diffc^{\mathrm{scalar}}$ are CLIP~\cite{clip}-encoded image condition and motion conditioning scalars for the 3D-UNet attention modules.
After the iterative denoising steps, the clean video latent $\diffz_0$ is decoded into an RGB video by the VAE decoder.
Although SVD can generate videos of promising quality, it lacks controllability over both scene and camera motion, restricting its use for scenarios such as view interpolation.

\subsection{Time-Reversal for video view interpolation}
\label{sec:timereversal}
To control the camera motion generation, Time-Reversal~\cite{timereversal} leverages a crucial property of SVD: the input conditioning image serves as the start frame of the generated video. 
Therefore, at each denoising step $\difft$, besides a denoising pass on the video latent $\diffz_{\difft}$ to get deonised $\diffz_{\difft-1}^{\mathrm{fwd}}$ with start-view image condition $\I^{\mathrm{start}}$,
Time-Reversal temporally reverses $\diffz_{\difft}$ 
as $\rev{\diffz}_{\difft}$ and 
denoises with another end-view image condition, $\I^{\mathrm{end}}$, to obtain the denoised $\rev{\diffz}_{\difft-1}^{\mathrm{end}}$. 
Subsequently, the two denoised $\diffz_{\difft-1}^{\mathrm{start}}$ and $\rev{\diffz}_{\difft-1}^{\mathrm{end}}$ are fused into $\diffz_{\difft-z}$ for the next denoising step $\difft-1$:
$
    \diffz_{\difft-1} = \trweight\diffz_{\difft-1}^{\mathrm{start}} + (1 - \trweight)\rev{\left( \rev{\diffz}_{\difft-1}^{\mathrm{end}} \right)}
$,
where $\trweight$ is the fusing weight and $\rev{\left( \rev{\diffz}_{\difft-1}^{\mathrm{end}} \right)}$ is the reverse of $\rev{\diffz}_{\difft-1}^{\mathrm{end}}$, back to the original video time space. 

By manipulating additional end-view image condition, $\I^{\mathrm{end}}$, Time-Reversal can perform video looping when the end view $\I^{\mathrm{end}}$ is identical to the start view $\I^{\mathrm{start}}$, and view interpolation between two different views of $\I^{\mathrm{end}}$ and $\I^{\mathrm{start}}$.
Nevertheless, a single view-interpolation video generation is insufficient for expanding an explorable 4D scene, especially SVD-generated video with 25 frames. Therefore, we incorporate multiple passes of Time-Reversal to acquire the scene motions in each part of a generated 3D scene (Sec~\ref{sec:video_gen}).

\begin{figure}
\centering
\includegraphics[trim={0cm 26.2cm 35.5cm 0.2cm},clip,width=0.96\linewidth]{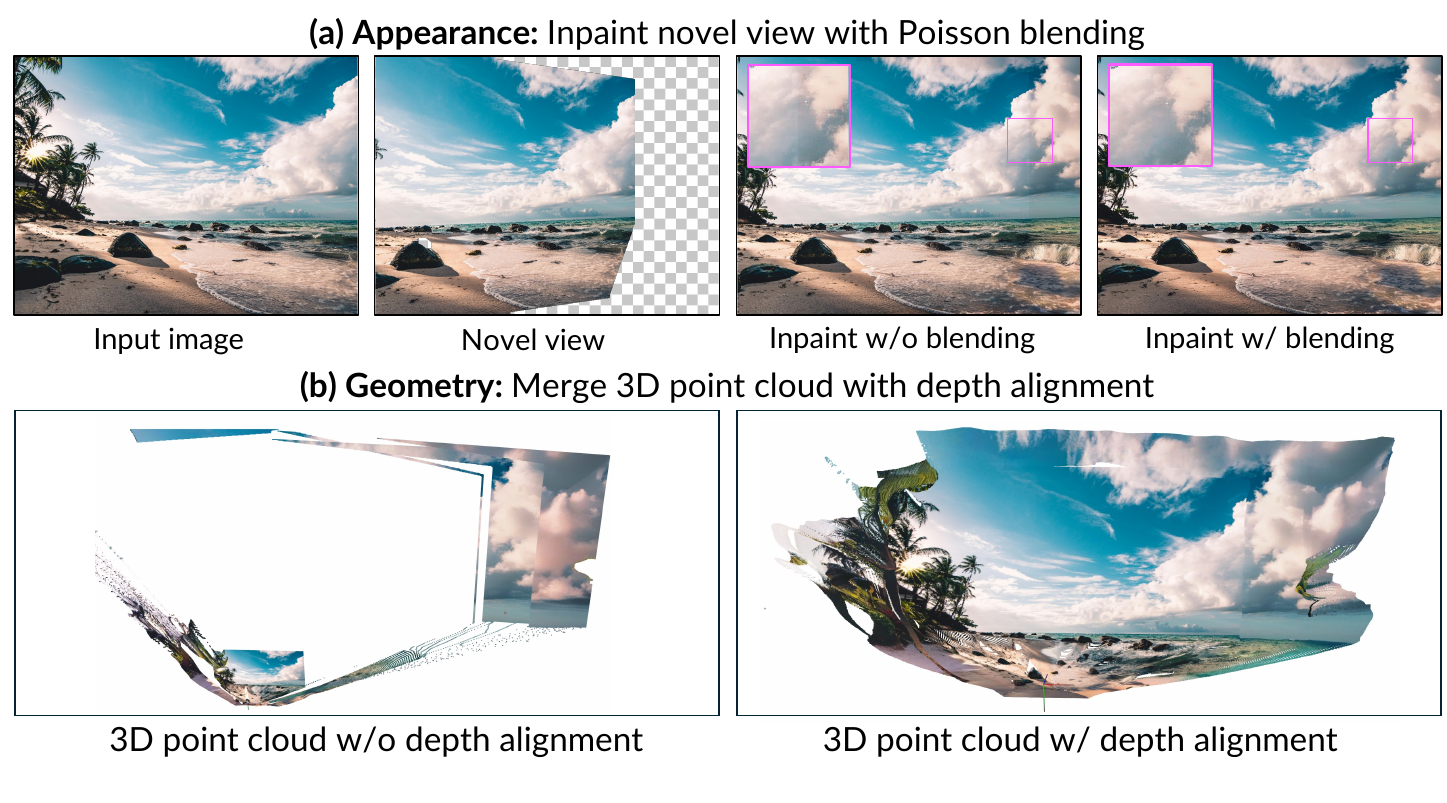}
\\
\vspace{-5.5mm}

\caption{\textbf{Handling appearance and geometry inconsistencies in 3D point cloud.} 
(a) The inpainting model~\cite{rombach2022high} may introduce color shifts caused by VAE, creating inconsistencies with the existing scene (zoomed-in). 
We mitigate these boundary seams using Poisson blending~\cite{Prez2003PoissonIE}.
(b) After merging the inpainted region with the existing point cloud, the newly estimated depth may be inconsistent with the existing geometry (left). 
To address this, we fine-tune the depth estimation model following SceneScape~\cite{fridman2024scenescape} to align the new depth with the known view depth. These steps ensure appearance and geometry consistency in each iteration of 3D point cloud generation.
}\vspace{-6mm}
\label{fig:3dpcd-gen}
\end{figure}

\subsection{4D Gaussian Splatting for 4D-scene view synthesis}
\label{sec:4dgs}
3D Gaussian Splatting (3DGS)~\cite{3dgs} recently has shown decent rendering quality and fast training and rendering speeds with an explicit representation, compared to NeRF~\cite{nerf} with neural implicit representation. 
The geometry of each Gaussian splat (GS) is represented by position $\gspos\in\R^3$ and an anisotropic covariance matrix $\Sigma$, where the covariance is decomposed by $\Sigma=\mathbf{R}\mathbf{S}\mathbf{S}^T\mathbf{R}^T$ with the rotation $\mathbf{R}\in\R^4$ and scaling $\mathbf{S}=\mathrm{Diag(s)}, s\in\R^3$. 
On the other hand, the appearance of a GS is parameterized by the color, $\gscolor$, encoded by spherical harmonics coefficients, and an opacity, $\a$. To render the color $\C$ of a given pixel at a view is rendered by depth-based sorting on the GS set $\{G_i\}_{i=1}^{N}$ and $\a$-blending with 2D-projected opacity:
\begin{equation}
\label{eq:render}\vspace{-2mm}
    \C = \sum_{i=1}^N \gscolor_i\a_i\prod_{j=1}^{i-1}(1-\a_j).
\end{equation}

To extend the 3DGS for dynamic scenes, 4DGS~\cite{4dgs} (4DGS) introduces a deformation module to deform each GS for dynamic motion. Given the 3D position $\gspos=(x, y, z)$ and the query time $t$, the deformation module obtains the feature $f$ using a spatiotemporal Hexplane encoder~\cite{kplanes}, $f$. Then, the feature $\hex(x, y, z, t)$ is fed into decoding MLPs $\gsmlp$ to acquire the deformation position $(\Delta x, \Delta y, \Delta z)$, rotation $\Delta r$, and scaling $\Delta s$ for the time $t$. With all GS $\{G_i\}_{i=1}^N$ deformed, the view of a dynamic scene is rendered by Eq.~\ref{eq:render}.

However, 4DGS is sensitive to noisy training views due to the reliance on accurate camera poses and wide coverage of viewing directions.
In our setting, which is based on multiple videos generated by SVD and Time-Reversal, the appearance and motion inconsistencies among videos can be harmful. 
To address this issue, we introduce per-video motion embedding and visibility mask to mitigate the blurriness caused by multi-video inconsistencies (Sec.~\ref{sec:train4d}).

\section{Method}
\label{sec:method}

Our three-stage pipeline for generating an explorable 4D scene with ambient motion is illustrated in Fig.\ref{fig:pipeline}.
In the following subsections, we provide detailed information about the three stages.

\subsection{3D scene generation}
\label{sec:3dscenegen}
Given a single-view point cloud $\pcd_0$ unprojected by the input image and the estimated depth, we expand the 3D point cloud through an iterative process of view extrapolations, consisting of novel-view inpainting and point-cloud merging. 
At each iteration $i$, we select a novel-view pose, $\pose_{i}$, to render the image $\I_{i}$ and depth $\D_i$ from the point cloud $\pcd_{i-1}$, with a mask $\m_{i}$ indicating the known regions.
The unseen regions in the rendered image are then filled by the inpainting model~\cite{rombach2022high} conditioned on image $\I_{i}$, inpainting mask $(1-\m_{i})$, and text $\prompt$.
We observe that the VAE encoding and decoding process produces color inconsistencies between the inpainted image $\I^{\mathrm{inpaint}}_{i}$ and the known regions in $\I_{i}$. To tackle this, we apply Poisson blending~\cite{Prez2003PoissonIE} on $\I^{\mathrm{inpaint}}_{i}$ with $\I_{i}$ along the mask borders to enhance color consistency (Fig.~\ref{fig:3dpcd-gen}a).
\begin{figure*}
\centering
\includegraphics[trim={0cm 11cm 26.5cm 0cm},clip,width=0.98\linewidth]{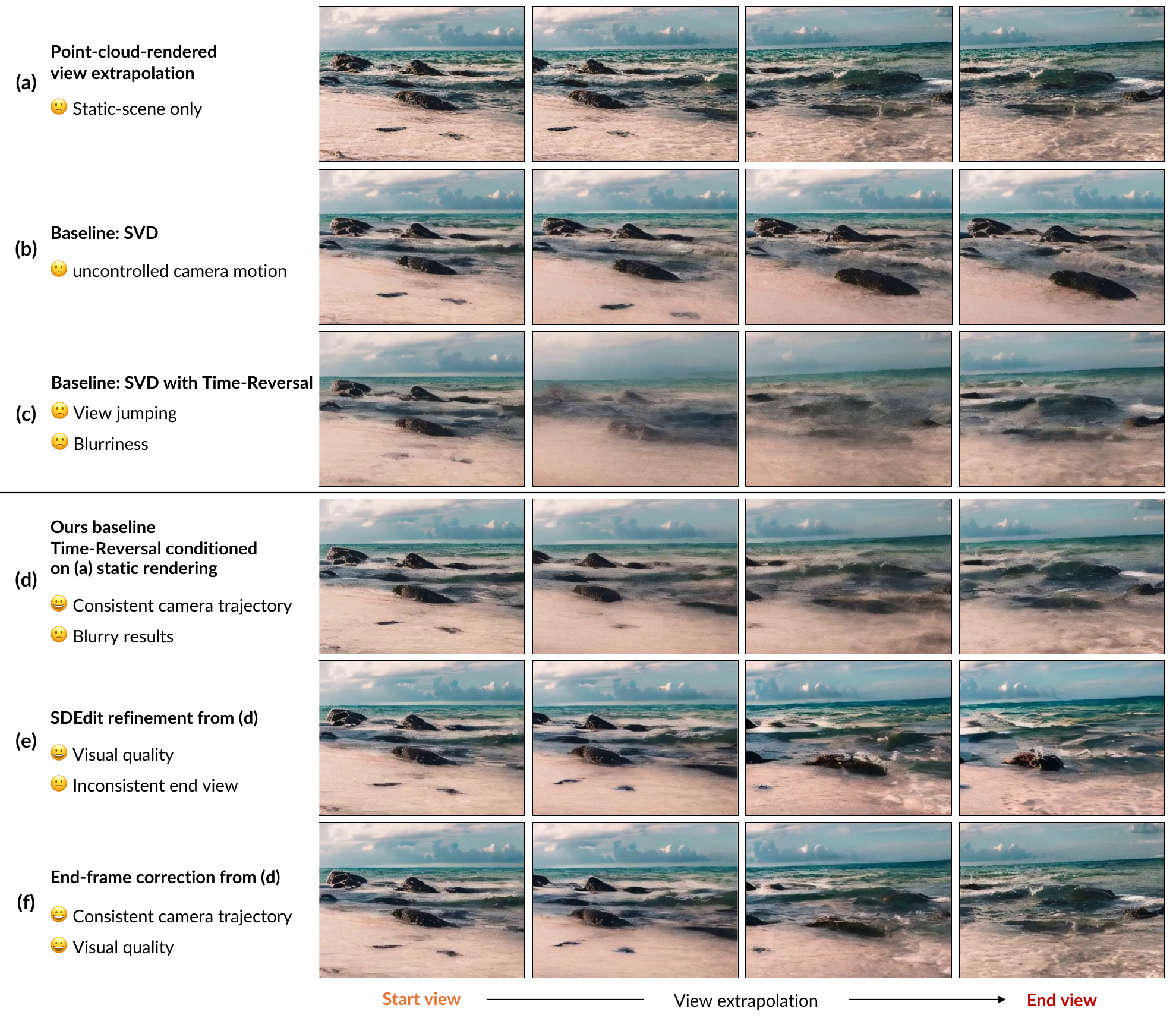}
\vspace{-5mm}
\caption{\textbf{Motion generation with controllable camera trajectory.} 
We aim to animate ambient scene dynamics while maintaining control over the camera trajectory. 
(a) We render a static scene video using the 3D point cloud with a smooth camera motion, serving as a condition for the animated videos.
(b) Naively applying SVD with only the start view results in uncontrollable camera poses.
(c) Time-Reversal ensures start and end view consistency but suffers from blurriness and camera trajectory deviations.
(d) Using the static scene video as a condition for SVD with Time-Reversal and SDEdit~\cite{sdedit} encourages following the desired trajectory but yields low-quality results.
(e) Applying SDEdit again with only the start view improves quality but causes camera pose deviations.
(f) We correct this by applying a smooth transition to the last frames using FILM~\cite{film} to match the end view. Our approach generates animated videos with ambient dynamics while respecting the specified camera trajectory.
}\vspace{-3mm}
\label{fig:video-gen}
\end{figure*}


\begin{figure}
\centering
\includegraphics[trim={0 28cm 31cm 0},clip,width=0.97\linewidth]{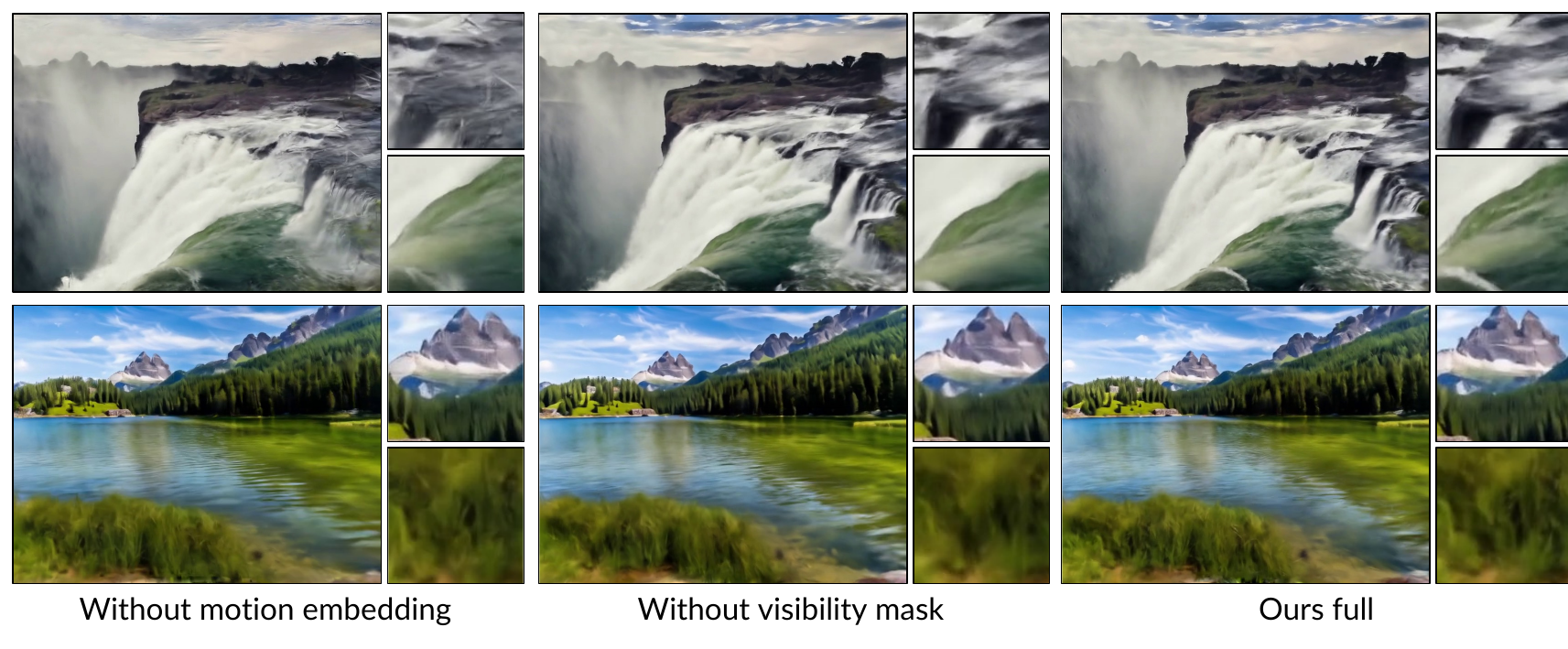}
\vspace{-5mm}
\caption{\textbf{Mitigating inconsistencies among multiple training videos.}
Training a 4D scene with independently animated videos inevitably leads to low-quality renderings. 
We introduce a per-video motion embedding into the deformation module to handle the inconsistency in the embedding space. 
The motion embedding results in a significant improvement in the quality (\eg the fluid motion).
Moreover, we apply visibility masks to the videos, ensuring that only one video observes each scene part, thereby mitigating the blurriness caused by multi-video inconsistencies.
}\vspace{-5mm}
\label{fig:train-4d}
\end{figure}
To unproject the inpainted image at novel view $\pose_i$, the depth $\D_i^\mathrm{inpaint}$ is estimated again by~\cite{depthanything}. 
Since the inconsistency between the new depth $\D_i^\mathrm{inpaint}$ and the known regions of existing depth $\D_i$ creates misalignment in 3D space (Fig.~\ref{fig:3dpcd-gen}b), we follow SceneScape~\cite{fridman2024scenescape} and align the depths by fine-tuning the last layer of the depth estimation model~\cite{depthanything} and minimizing the disparity between the two depths, $\D_i^\mathrm{inpaint}$ and $\D_i$, in the known region $\m_i$:
\begin{equation}
\label{eq:pcd_depth_align}
    \Loss^{\mathrm{inpaint}}_{\mathrm{depth}} = \sum \left| \frac{{\D_i^\mathrm{inpaint}}^{-1} - {\D_i}^{-1}}{{\D_i^\mathrm{inpaint}}^{-1} + {\D_i}^{-1}} \right|^2_2 \odot \m_i,
\end{equation}
where $\odot$ is the element-wise multiplication.
We then use the aligned depth $\D^{*\mathrm{inpaint}}_i$ to lift the inpainted region $(1-\m_i)$ of image $\I_i^\mathrm{inpaint}$ into 3D to form $\pcd_i^\mathrm{inpaint}$, and merge it with the existing one $\pcd_{i-1}$ into $\pcd_{i}=\pcd_{i-1} \cup \pcd_{i}^{\mathrm{inpaint}}$. 
By iterating this process, an expanded 3D static point cloud $\pcd$ is used for view exploration.

\subsection{Multi-video animation}
\label{sec:video_gen}
In Stage~2, we utilize the image-to-video generator~\cite{svd} to generate ambient motion using the static 3D scene $\pcd$ from Sec.~\ref{sec:video_gen}.
To bake the motion into a 4D scene representation for view exploration, controlling the camera motions of the generated videos is essential for the 4D scene optimization in Stage 3 (Sec.~\ref{sec:train4d}).
Therefore, instead of a single-image condition in the original SVD, we condition the video generation process with a static-scene video rendered by a preset camera trajectory from the point cloud $\pcd$.
Since a single video cannot encompass the entire 3D scene, we acquire a set of videos $\{\V^{\mathrm{static}}_k\}_{k=1}^K$ to ensure complete view coverage by rendering $K$ videos with camera trajectories $\{\pose^{\mathrm{start}}_k, ..., \pose_k^{\mathrm{end}}\}_{k=1}^K$, 
where each video $\V^{\mathrm{static}}_k = \{\I_k^{\mathrm{start}}, ..., \I_k^{\mathrm{end}}\}$ captures a part of the scene. We omit $k$ here for simplicity.
In practice, we take the \textbf{view-extrapolation paths} from Stage~1 since the extrapolation paths for 3D scene generation will cover the entire scene.
For each path, we interpolate between the known view $\pose_{i-1}$ and the novel view $\pose_{i}$ to form a camera trajectory of length $T$, where $T$ is the temporal resolution of the SVD model. 
This static-scene extrapolation video serves as the condition of the video generation (Fig.~\ref{fig:video-gen}a) to obtain the ambient scene motion while retaining its camera trajectory. 

However, the original SVD cannot control camera motion in the generated videos (Fig.~\ref{fig:video-gen}b).
Although some methods~\cite{wang2023motionctrl,he2024cameractrl} extend the pretrained video diffusion models by inserting additional motion conditioning modules, the generated videos may still not follow the camera condition~\cite{wang2023motionctrl} or contain insufficient ambient scene motion~\cite{he2024cameractrl}. 
Alternatively, Time-Reversal~\cite{timereversal} builds upon SVD with two end-view image conditions, $\I^{\mathrm{start}}, \I^{\mathrm{end}}$, to allow better control over camera motion and scene motion for video looping and view interpolation (Sec.~\ref{sec:timereversal}). 
However, the two end-view conditions do not guarantee the camera motion in the middle frames, so it may still deviate from the specified trajectory (Fig.~\ref{fig:video-gen}c).

To improve camera control, we follow SDEdit~\cite{sdedit} and generate the ambient motion conditioned on the static-scene rendering $\V^\mathrm{static}=\{\I^{\mathrm{start}}, ..., \I^{\mathrm{end}}\}$. 
Concretely, we encode the static-scene video into the latent $\diffz_0^{\mathrm{static}}$ and then add noise to a diffusion step $\trstep$ in the diffusion noise schedule as a perturbed latent $\diffz_\trstep^{\mathrm{static}}$. 
Therefore, the denoising process of Time-Reversal is performed on the perturbed $\diffz_\trstep^{\mathrm{static}}$ from the step $\trstep$ to get a clean video latent $\diffz_0^{\mathrm{dyn}}$ with dynamic motion. 
The VAE-decoded RGB video $\V^{\mathrm{dyn}}=\{\I_1, ..., \I_T\}$ then could have scene motion at the temporal resolution $T$ and better camera motion consistency with the specified trajectory (Fig.~\ref{fig:video-gen}d). 

Nonetheless, we found that Time-Reversal still yields low visual quality due to the noise-averaging steps. 
To restore visual quality, we apply a second round of SDEdit to the initial animated video from the denoising step $\refinestep$ with SVD only. 
Although the details can be enhanced, the appearance and camera motion in the later frames may slightly deviate from the specified static-rendering video without the end-view condition from Time Reversal (Fig.~\ref{fig:video-gen}e). To address this, we adopt FILM~\cite{film}, to replace the last $n$ frames of the video, $\{\I_{T-n+1}, ..., \I_T\}$, with $\{\I^{\mathrm{FILM}}_{T-n+1},..., \I^{\mathrm{FILM}}_{T-1}, \I^{\mathrm{end}}\}$, creating a smooth transition to the conditioned end-view $\I^{\mathrm{end}}$, where $\{\I^{\mathrm{FILM}}_{T-n+j}\}_{j=1}^{n-1}$ is the frame interpolation output with two fixed view $\I^{\mathrm{FILM}}_{T-n}$ and $\I^{\mathrm{end}}$. Consequently, we can obtain a set of animated videos, $\{\V_k^{\mathrm{dyn}}\}_{k=1}^K$, with specified camera trajectories and desired scene motion.

\subsection{4D-scene training}
\label{sec:train4d}
With the expanded 3D point cloud $\pcd$ and multiple animated videos $\{\V^{dyn}_k\}_{k=1}^K$ covering the 3D scene, we now fit the 4D scene. The canonical 3DGS is initialized by the point cloud $\pcd$. For each video $\V^{\mathrm{dyn}}_k$, we assign timestamps $t=\{1, 2, ..., T\}$ for the $T$ frames.

The 4DGS optimization is sensitive to the quality of training data. In contrast to the original 4DGS~\cite{4dgs} trained on single or multiple synchronized videos, the multi-view videos, in our case, contain significant inconsistencies in appearance and motion by separate passes of video generations.
The inconsistencies could harm the optimization of 4DGS, leading to noticeable blurriness and floaters. To address this, we introduce \textbf{per-video motion embeddings} and \textbf{visibility masks} to the training in both implicit and explicit approaches to handle the multi-video inconsistencies.

Inspired by NeRF-W~\cite{nerfw} for handling inconsistent samples in an embedding space, we employ a \emph{per-video} motion embedding, ${\embed_k}_{k=1}^K\in\R^W$, instead of per-image embedding, to ensure consistent motion within each video while managing inconsistencies across videos. Specifically, given a 3D position $\gspos=(x, y, z)$ for each Gaussian splat (GS) and the query time $t$, the deformation module first obtains the spatiotemporal feature, $\hex(x,y,z,t)$, using a Hexplane encoder $\hex$ (as detailed in Sec.~\ref{sec:4dgs}). We concatenate this feature with the motion embedding $\embed_k$ and feed into MLP decoder $\gsmlp$ to obtain video-dependent deformation terms for training:
$
    \{\Delta x_k, \Delta y_k, \Delta z_k, \Delta r_k, \Delta s_k\} = \gsmlp(\hex(x,y,z,t) \oplus \embed_k)
$,
where $\oplus$ denotes concatenation.  
After training, all embeddings are averaged into a global motion embedding to render global consistent motion.
To regularize the embeddings, we constrain the embedding norm in a 1-norm space~\cite{feng2022viinter}. 
In practice, we set the embedding dimension $W=16$.

Implicitly handling inconsistencies in the motion embedding space may still be insufficient, so we further apply an explicit visibility mask $\{\M_k\}_{k=1}^K$ to each video.
Concretely, the visibility masks ensure that each part of the scene is observed by only one video (as shown in Fig.~\ref{fig:pipeline}). 
We jointly consider camera poses and the 3D static point cloud to determine the visibility masks. 
The length of a viewing ray, $\|\ray_k^j\|$, can be obtained by the depth of the 3D point cloud. The view-angle score $\vec{\ray}_k^j$ of the viewing ray $\ray^j_k$ w.r.t to the camera-looking direction, $\lookat^j_k$, is computed by $\ray^j_k\cdot\lookat_k^j / \|\ray^j_k\| \|\lookat_k^j\|$.
The overall viewing length $\|\ray_k\|$ and view-angle score $\vec{\ray}_k$ of a video $\V_k$ is the average over all the frames.

\begin{figure*}
\centering
\includegraphics[trim={0cm 18.5cm 32.3cm 0cm},clip,width=\linewidth]{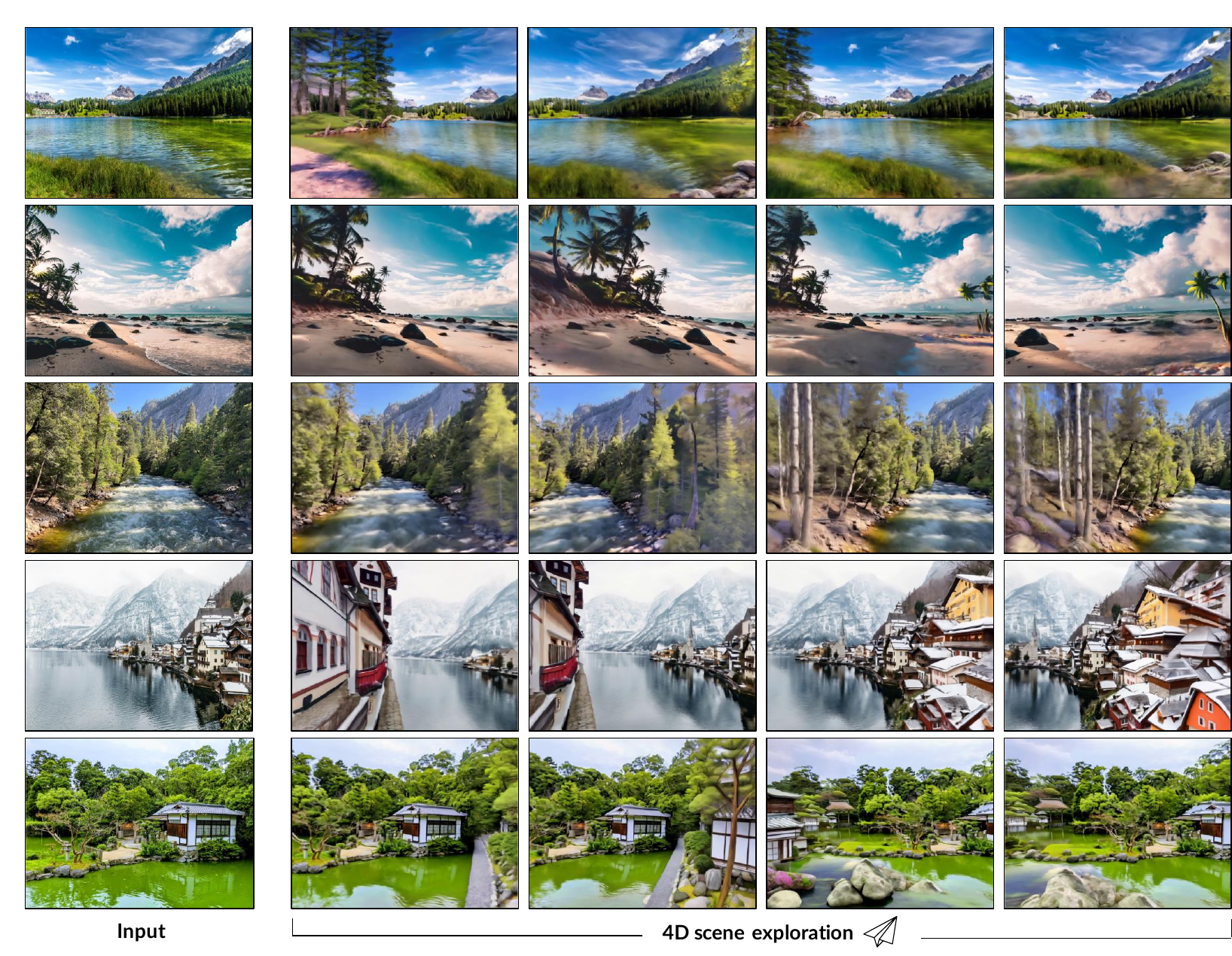}
\vspace{-6.5mm}
\caption{\textbf{Qualitative results of real photo inputs. Our method turns a real static image into an alive 4D scene with ambient scene motions.}
}
\label{fig:real-ex}
\end{figure*}

We then use the two metrics to find the best video that views the scene part with close and forward-facing views: $k^*=\arg\max_k \frac{1}{\|\ray_k\|} + \beta\vec{\ray}_k$,
where $\beta (=1)$ is a scalar to balance the two terms. The visibility masks for all videos are then acquired by the $k^*$ for each scene part. The hard selection ensures each part is only trained by one video to avoid blurriness, but we also expand the visibility mask by a small width with a soft decreasing weight to encourage a smooth spatial transition among the training videos. Note that the visibility masking may result in separate motions because different videos train different areas. While we focus on generating a scene with ambient motion, the separate motion may be negligible. Then, the visibility masks are used as weighting maps to compute the RGB and depth loss during training:
    $\Loss_{\mathrm{rgb}} = \sum \|\tilde\I_k - \I_k\|_1 \odot \M_k$, and
    $\Loss_{\mathrm{depth}} = \sum |\tilde{\D}_k^{-1} - \D_k^{-1}|_1 \odot \M_k$,
where the depth loss is slightly different from Eq.~\ref{eq:pcd_depth_align}. 
Here, we apply more penalty to near scenes and more tolerance to the error in far scenes since the point cloud depth alignment may still exist misalignment in the far scenes. 
Notably, the 4D scene depth is trained by the depth of the 3D static point cloud. It may not be suitable when the scene contains outstanding moving objects, such as humans and animals. Since we are generating a 4D scene with ambient motions, we found it sufficient to supervise learning 3D scene structure. During 4DGS fitting, to prevent quality degradation from training with the VAE-decoded videos via SVD, we also use the point-cloud-rendered images in higher quality to supervise the canonical 3DGS model. We adopt the rigidity loss~\cite{luiten2023dynamic} $\Loss_\mathrm{rigidity}$ to encourage neighboring Gaussian splats to have a similar deformation for propagating the high-quality canonical appearance to 4D rendering. The total is then computed by
    $\Loss = \Loss_{\mathrm{rgb}} + \lambda_{\mathrm{depth}}\Loss_{\mathrm{depth}} + \lambda_{\mathrm{rigidity}}\Loss_{\mathrm{rigidity}},$
where $\lambda_{\mathrm{depth}}$ and $\lambda_{\mathrm{rigidity}}$ balance the loss terms. Finally, the 4D scene optimization learns the motion from multi-video with proper handling of multi-video inconsistencies, which results in a 4D explorable scene with ambient motion and sharp details (Fig.~\ref{fig:train-4d}).

\begin{figure*}
\centering
\includegraphics[trim={0cm 15cm 32.3cm 0cm},clip,width=\linewidth]{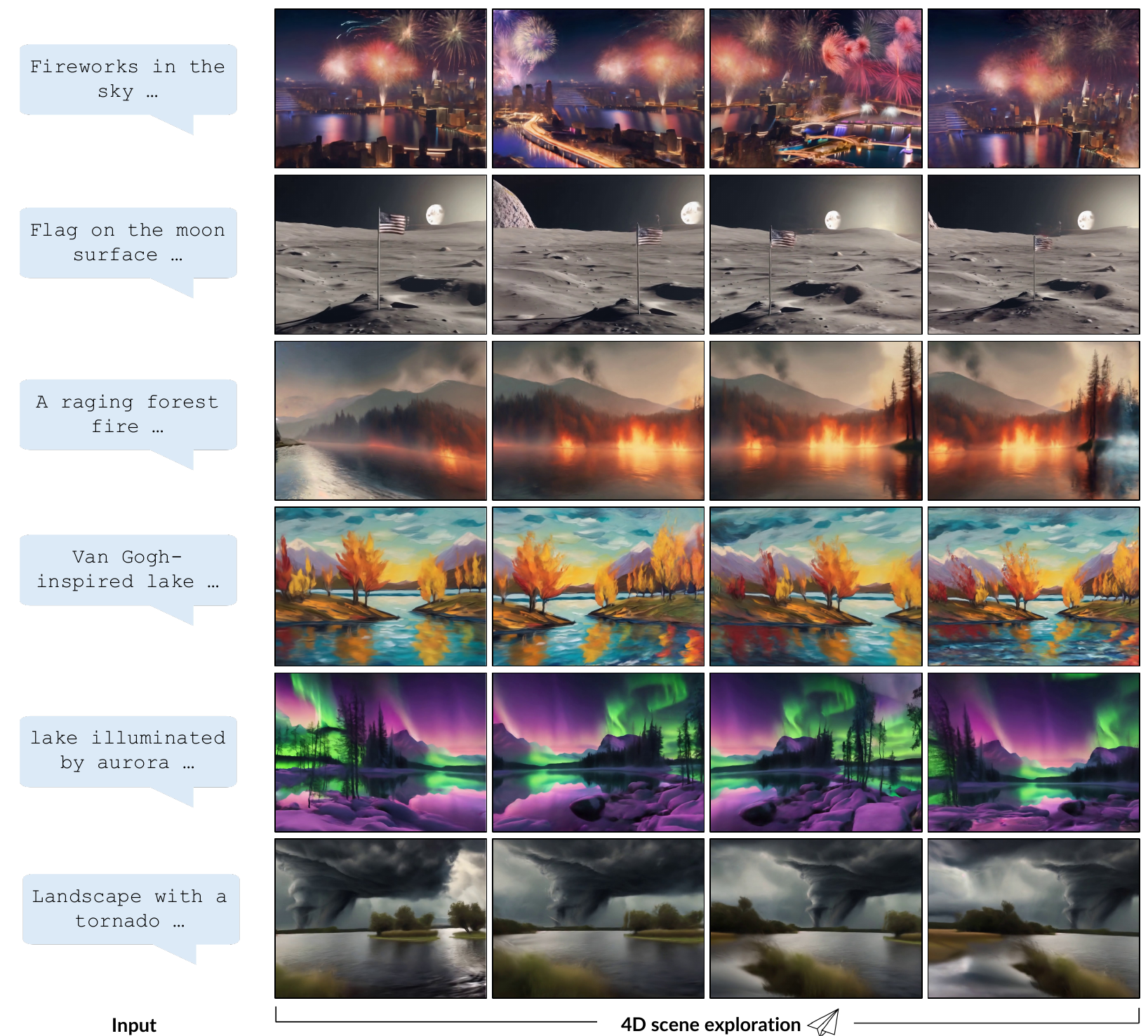}
\vspace{-6.5mm}
\caption{\textbf{Qualitative results of text inputs.} Our method can generate diverse ambient scene motion in various text-guided generated scenarios.
}
\label{fig:text-ex}
\end{figure*}
\begin{figure*}
\centering
\includegraphics[trim={0cm 34cm 28.5cm 0cm},clip,width=\linewidth]{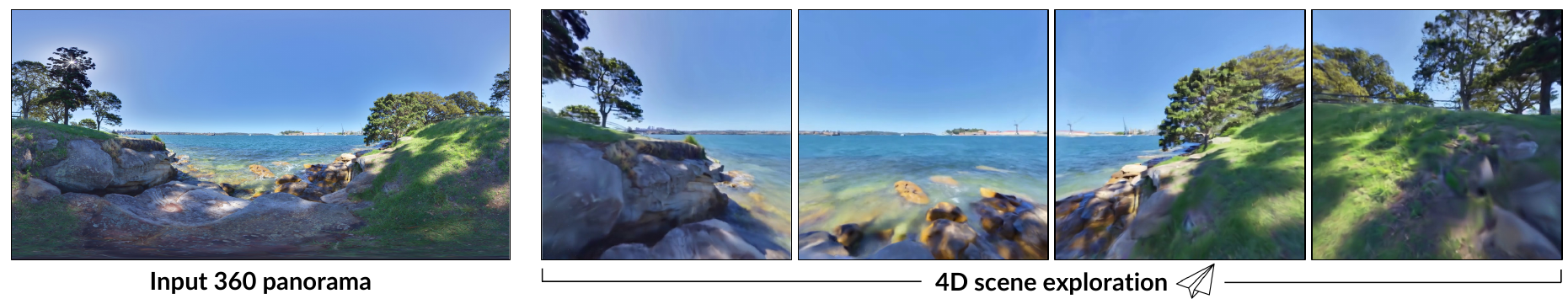}
\vspace{-6.5mm}
\caption{\textbf{Generating immersive 4D scene from 360 panorama image.} Our method can also take a 360 panorama image to generate an explorable 4D scene with ambient motion.
}
\label{fig:pano-ex}
\end{figure*}
\begin{figure*}
\centering
\includegraphics[trim={0cm 32cm 5.5cm 0cm},clip,width=\linewidth]{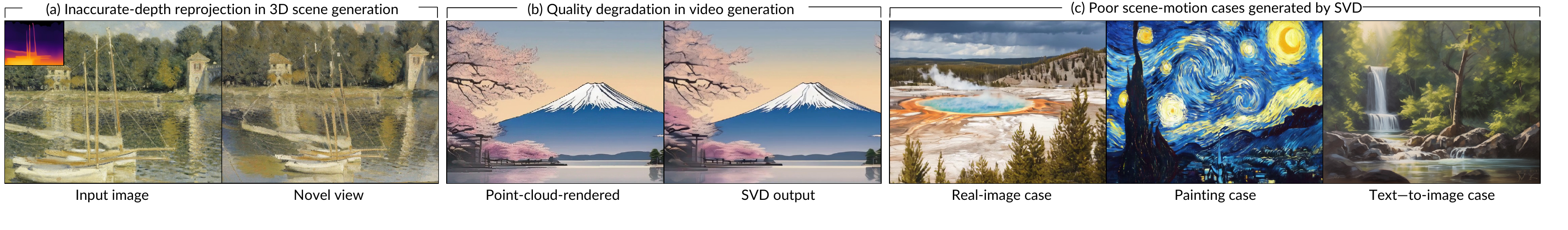}
\vspace{-9mm}
\caption{\textbf{Limitations.} 
Our method relies on a series of successes in 3D scene generation and video generation. 
(a) In 3D scene generation, imperfect depth estimation/reprojection often occurs in thin structures, leading to noisy results. (b) In video generation, the quality may degrade through the encoding and decoding process of SVD VAE, causing a blurry 4D scene. 
(c) The scene motion generation is difficult to control. 
SVD often fails to generate plausible motion for paintings and text-conditioned generated images. 
We expect more advanced video generation models may achieve better quality and controllability, but most of these are not available to the public.
}
\label{fig:failure}
\end{figure*}
\section{Experimental Results}


\topic{Implementation details.}
In Stage 1, for 3D scene generation, we typically expand the scene ten times, taking about 15 minutes. In Stage 2, for multi-video generation, we create 10 extrapolation videos ($K=10$). The animation process, including quality refinement and end-frame correction, takes roughly 7 minutes per video with $\trstep=16$ for Time-Reversal and $\refinestep=9$ for SVD refinement. 
We re-implement Time-Reversal~\cite{timereversal} by ourselves.
In Stage 3, the canonical 3DGS is trained for 3,000 iterations with a maximum spherical harmonics order of 3. Then, the 4DGS is further trained for 15,000 iterations, with $\lambda_{\mathrm{depth}}=1$ and $\lambda_{\mathrm{rigidity}}=1$, taking around 1.5 hours. Overall, the 4D scene generation process takes about 3.5 hours. Notably, the baseline method, which involves reconstructing a single SVD video using depth and pose estimation\cite{casualsam} and then training with 4DGS, requires more than 4 hours.

\subsection{Qualitative results}
As 4D scene generation is a new problem without an existing benchmark, we showcase a free-view exploration of the generated 4D scenes.
In Fig.~\ref{fig:real-ex}, our method expands the real input image into a larger scene with ambient motions.
In Fig.~\ref{fig:text-ex}, we demonstrate diverse text-guided 4D scene generation scenarios with various scene ambient motion. 
Furthermore, our method can also turn a 360 panorama image into a 4D immersive scene (Fig.~\ref{fig:pano-ex}), which can be further applied to VR applications. 

\subsection{Human perceptual evaluation}

The primary application of our work is for creative and entertainment purposes. 
It's difficult to evaluate such generated results with standard reference-based metrics.
We thus employ human perceptual evaluation on the generated 4D-scene videos, focusing on detail quality, visual appeal, and view explorability. We compare our method against a baseline and conduct an ablation study on each component. Participants answered binary choice questions like “Which video has better overall visual appeal?” and we collected feedback from 32 users. As shown in Table~\ref{tab:user_study}, our method surpasses the baseline in visual appeal and view explorability. Our lower preference ratio for detail quality is due to inconsistencies from multi-video training, while baseline is trained with single-video which captures more consistent details yet with limited viewpoints.
For the ablation study, we compare against our ablated methods without SDEdit quality refinement in the video generation stage, per-video motion embedding, and visibility masking in the training stage to handle the multi-video inconsistencies.
The results indicate a general preference for our full method over the ablated versions. The improvement from visibility masking was less significant, likely because detail improvements were not easily noticeable in videos with rapid view changes. However, quality improvements are evident in the image comparison shown in Fig.~\ref{fig:train-4d}.

\begin{table}[]
    \caption{Human perceptual user study for comparison with baseline and ablation study on details, visual appeal, and explorability.}
    \vspace{-3mm}
    \resizebox{\linewidth}{!}{\begin{tabular}{l|c c c}
    \toprule
      Questions.   & Detail.  & Appeal. & Explore.\\
      \midrule
      Ours over baseline. & 61.5\% & 82.3\% & 91.7\%\\ 
      \midrule
      Ours over w/o SDEdit quality refine. & 88.2\% & 80.6\% & -\\
      Ours over w/o motion embed. & 59.8\% & 65.9\% & - \\
      Ours over w/o mask. & 50.5\% & 52.2\% & - \\ 
    \bottomrule
    \end{tabular}}\vspace{-5mm}
    \label{tab:user_study}
\end{table}

\section{Limitation and Discussion}
We acknowledge that the quality of our method is far from perfect. 
Our method is complex and time-consuming with multiple stages, and relies on a series of successful processes to generate a 4D scene with plausible ambient motion. 
In the 3D scene generation stage, inaccurate depth estimation can be revealed in a distant novel view with a large viewpoint change, restricting the 3D scene expansion.
In addition, the depth estimation model often fails to estimate thin structures, leading to distorted results in the novel view (Fig.~\ref{fig:failure}a). 
Furthermore, the noisy reprojection will also lead to undesirable video generation in Stage~2.
In the multi-video generation stage, SVD significantly degrades the input image quality (Fig.~\ref{fig:failure}b), causing a blurry 4D scene reconstruction in Stage 3.
Besides, SVD still lacks scene motion control and often fails in non-real image cases, such as paintings, and diffusion-generated images (Fig.~\ref{fig:failure}c).
As mentioned by Time-Reversal~\cite{timereversal}, the SVD only takes a few motion scalars as animation conditions, which are ambiguous and often require careful tuning on the new input. 
Despite these major limitations, our method is the first to enable the generation of explorable 4D scenes with ambient motions. When applied to a more reliable video generator in the future, we believe that our proposed pipeline can significantly enhance 4D scene generation, leading to improved quality and robustness.
{
    \small
    \bibliographystyle{ieeenat_fullname}
    \bibliography{ms}
}

\end{document}